\documentclass{article}
\usepackage{amsmath,amssymb,amsfonts}
\usepackage{algorithmic}
\usepackage{authblk}
\usepackage{float}
\usepackage{booktabs}
\usepackage{multirow}
\usepackage{graphicx}
\usepackage{microtype}
\DeclareMathOperator*{\argmax}{argmax}
\begin{document}

\providecommand{\keywords}[1]
{
  \small	
  \textbf{\textit{Keywords---}} #1
}
\title{Gradually %Attention AE/ME. The following layout issues have not been checked by the English Editing Department and must be carefully verified by the AE/Layout Department: All callout issues, bold usage of callouts, and references to callouts in the text. Correct callout usage in figures. Figure and Table layout issues. Footnote formatting and Glossaries have not been checked. En dash usage for negative values, en dash usage to indicate relationships, en dash usage to indicate bonds (especially in chemistry). The English Editing Department is not responsible for correct italic usage for genes, proteins and technical terminology. This responsibility belongs to the authors. The following are also not checked: spacing between numbers and units of measurement, ratios, en dashes for ranges, date and time formats, punctuation in equation lines, and less than/more than spacing (< >). Finally, capitalization and layout of titles/headings must be properly checked as well as ensuring 'Eq.' and 'Fig.' are properly spelled out, as these are layout issues.
  Applying Weakly Supervised and Active Learning for Mass Detection in Breast Ultrasound~Images}
\maketitle

% Author Orchid ID: enter ID or remove command
% \newcommand{\orcidauthorA}{0000-0000-000-000X} % Add \orcidA{} behind the author's name
%\newcommand{\orcidauthorB}{0000-0000-000-000X} % Add \orcidB{} behind the author's name

% Author homepage: enter homapage URL or remove command
%\newcommand{\homepageauthorA}{https://www.mdpi.com/} % Add \homepageA{} behind the author's name
%\newcommand{\homepageauthorB}{https://www.mdpi.com/} % Add \homepageB{} behind the author's name

% Authors, for the paper (add full first names)
% \author[1]{JooYeol Yun}
% \author[2]{JungWoo Oh}
% \author[3]{IlDong Yun}
% \affil[1]{Department of Computer Science and Engineering, Korea University, Seoul, 02841 South Korea}
% \affil[2]{Department of Computer Science and Engineering, Korea University, Seoul, 02841 South Korea}
% \affil[1]{Department of Computer and Electronic Systems Engineering, Hankuk University of Foreign Studies, Yongin, 17035 South Korea}
\author{JooYeol Yun$^{1}$, JungWoo Oh$^{1}$, IlDong Yun$^{2}$  \\
        \small $^{1}$Korea University \\
        \small $^{2}$Hankik University of Foreign Studies
}
% Current address and/or shared authorship
% \firstnote{Current address: Affiliation 3} 
% \secondnote{These authors contributed equally to this work.}
% The commands \thirdnote{} till \eighthnote{} are available for further notes

%\simplesumm{} % Simple summary

%\conference{} % An extended version of a conference paper

% Abstract (Do not insert blank lines, i.e. \\) 
\begin{abstract}
We propose a method for effectively utilizing weakly annotated image data in an object detection tasks of breast ultrasound images.
Given the problem setting where a small, strongly annotated dataset and a large, 
weakly annotated dataset with no bounding box information are available, 
training an object detection model becomes a non-trivial problem.
We suggest a controlled weight for handling the effect of weakly annotated images in a two stage object detection model.
We~also present a subsequent active learning scheme for safely assigning weakly annotated images a strong annotation using the trained model.
Experimental results showed a 24\% point increase in correct localization (CorLoc) measure, which is the ratio of correctly localized and classified images, 
by assigning the properly controlled weight.
Performing active learning after a model is trained showed an additional increase in CorLoc.
We tested the proposed method on the Stanford Dog datasets to assure that it can be applied to general cases, 
where strong annotations are insufficient to obtain resembling results.
The presented method showed that higher performance is achievable with lesser annotation effort.
\end{abstract}

% Keywords

% The fields PACS, MSC, and JEL may be left empty or commented out if not applicable
%\PACS{J0101}
%\MSC{}
%\JEL{}

%%%%%%%%%%%%%%%%%%%%%%%%%%%%%%%%%%%%%%%%%%
% Only for the journal Diversity
%\LSID{\url{http://}}

%%%%%%%%%%%%%%%%%%%%%%%%%%%%%%%%%%%%%%%%%%
% Only for the journal Applied Sciences:
%\featuredapplication{Authors are encouraged to provide a concise description of the specific application or a potential application of the work. This section is not mandatory.}
%%%%%%%%%%%%%%%%%%%%%%%%%%%%%%%%%%%%%%%%%%

%%%%%%%%%%%%%%%%%%%%%%%%%%%%%%%%%%%%%%%%%%
% Only for the journal Data:
%\dataset{DOI number or link to the deposited data set in cases where the data set is published or set to be published separately. If the data set is submitted and will be published as a supplement to this paper in the journal Data, this field will be filled by the editors of the journal. In this case, please make sure to submit the data set as a supplement when entering your manuscript into our manuscript editorial system.}

%\datasetlicense{license under which the data set is made available (CC0, CC-BY, CC-BY-SA, CC-BY-NC, etc.)}

%%%%%%%%%%%%%%%%%%%%%%%%%%%%%%%%%%%%%%%%%%
% Only for the journal Toxins
%\keycontribution{The breakthroughs or highlights of the manuscript. Authors can write one or two sentences to describe the most important part of the paper.}

%\setcounter{secnumdepth}{4}
%%%%%%%%%%%%%%%%%%%%%%%%%%%%%%%%%%%%%%%%%%
% \begin{document}
%%%%%%%%%%%%%%%%%%%%%%%%%%%%%%%%%%%%%%%%%%

%%%%%%%%%%%%%%%%%%%%%%%%%%%%%%%%%%%%%%%%%%
% \setcounter{section}{-1} %% Remove this when starting to work on the template.
\section{Introduction}
% The introduction should briefly place the study in a broad context and highlight why it is important. It should define the purpose of the work and its significance. The current state of the research field should be reviewed carefully and key publications cited. Please highlight controversial and diverging hypotheses when necessary. Finally, briefly mention the main aim of the work and highlight the principal conclusions. As far as possible, please keep the introduction comprehensible to scientists outside your particular field of research. Citing a journal paper~\cite{ref-journal}. And now citing a book reference~\cite{ref-book}. Please use the command \citep{ref-journal} for the following MDPI journals, which use author-date citation: Administrative Sciences, Arts, Econometrics, Economies, Genealogy, Humanities, IJFS, JRFM, Languages, Laws, Religions, Risks, Social Sciences.
\label{section:introduction}
Breast cancer is the second leading cause of death for women all over the world, while their cause still remains unknown~\cite{i11}.
Like most cancer, early detection plays an important role in reducing the death rate~\cite{i12}.
While digital mammography is the most commonly used technique for detecting breast cancer, its limitations are clear when observing dense breasts, where lesions can be hidden by tissues having similar attenuation~\cite{i13}.
Ultrasound imaging is a complementary method for digital mammography, due to its sensitivity, cost-effectiveness, and safety.
However, analyzing ultrasound images is not a straight forward task due to the presence of noise and, thus, requires a skilled radiologist.
Computer Aided Diagnosis (CAD) could reduce the dependency on the radiologist and also be beneficial for detecting breast cancer~\cite{i14}.

Breast ultrasound (BUS) images follow the characteristics of a typical ultrasound image, which is generally low in resolution and containing noise.
Resolution can be enhanced using higher frequency waves, which will on the other hand limit the penetration depth~\cite{i15}.
While the usual process of diagnosing a BUS image will accompany clinical palpation when palpable, CAD would only have the BUS image available~\cite{i16}.
Additionally, BUS images are not taken at a fixed angle during diagnosis, unless special superimposing among all aspect angles will be performed later~\cite{i17}.
The loss of palpation information and the diversity of image aspects makes it challenging for a CAD to~improve.

Conventional methods for BUS image classification that does not use a neural network framework are based on preprocessing and feature extraction of BUS images following region detection or segmentation with those features.
Selected features after region detection are classified with different methods~\cite{i11}.
Most of these works focus on feature extraction of the image when the aim is to classify an image to benign or malignant.
Other works that aim to localize the lesion use rule based approaches, such as the deformable parts model~\cite{i22}.

Recent deep learning based frameworks conduct both classification and region detection as annotated data became more available.
Semantic segmentation is performed with BUS images in~\cite{i23} by replacing the last three fully connected layers of AlexNet to fully convolutional networks that perform pixel-wise classification.
The work utilizes mask labels that have labels for every pixel in the image as ground truth for all of the images.
Mask annotations require more labor from a clinician and, therefore, are harder to obtain.
Shin~et~al. \cite{i24} proposed a method for object localization and classification using a Faster-RCNN model.
While using bounding box annotations as ground truth for the localization task, it makes use of weakly supervised data only comprised of image level label to aid the classification~model.

We present a method for sequentially localizing and classifying BUS images based on the Faster-RCNN model presented in~\cite{i25}.
We train a convolutional neural network (CNN) for bounding box regression and mass classification.
A fully connected network (FCN) that classifies bounding boxes as either benign, malignant, or~background is trained concurrently with the earlier network.
The ground truth information are bounding box coordinates and classification labels for each mass.
However, BUS data consisting of only the classification labels for each image are more accessible while bounding box annotations require additional expert effort.
As BUS image classification still remains a difficult problem, a large dataset size will be beneficial to enhance the~performance.

Weakly supervised learning is a technique for machine learning with noisy, sparse annotations.
A~customized alteration, depending on the degree of the annotations, is needed in order to use data with different levels of supervision.
Methods for utilizing image level annotations for segmentation are proposed in~\cite{i31}.
An initial segmentation model is trained using a few strongly annotated images.
Images with no mask annotations are given a pseudo mask ground truth generated by the initially trained model and the second model is trained to perform both segmentation and image level classification with these pseudo annotations.
Generative adversarial networks (GANs) are tuned to perform semantic segmentation while using both image level annotations and generate mask annotations.
Shin~et~al. \cite{i24} uses both bounding box annotations and image level labels to localize and classify objects using multiple-instance learning (MIL).
Images without bounding box annotations are given a bounding box chosen from a bag of bounding boxes presented during the localization stage.
Various methods for choosing an object among the candidates are~tested.

Active learning is a mechanism for expanding the given dataset by labeling unlabeled data with the train model.
User intervention for labeling is encouraged during the whole training process.
Active learning can be applied to different types of datasets and fields where data is scarce.
Mask prediction for lung CT images generated by unsupervised segmentation is used as ground truth annotation for training a supervised segmentation network~\cite{i34}.
The segmentation network is trained multiple times while using the mask prediction from the previous model as the ground truth, progressively improving after each training~session.

We propose an appropriate method for controlling the influence of weakly labeled data in a Faster-RCNN based object detection model.
The presented method shows increase in correct localization (CorLoc) measures, which is preferred over mean average precision (mAP) in medical imaging, 
and fraction of lesions detected, which measures the localization performance.
The presented method assumes a relatively small strongly annotated dataset insufficient for achieving high classification capability and a larger dataset with weakly labeled images, which is a typical setting for medical imaging where making strong annotations are~costly.

The main contributions of this work are, first,
designing a reasonable method of controlling the effect of weakly labeled data in an end-to-end object detection model and, second, designing an acceptable approach for actively assigning annotations for weakly labeled data, supplementing the insufficient annotations for object detection.
The strongly annotated data, $D_{strong}$, contain a single bounding box coordinate and the box classification label per image, and the weakly labeled data, $D_{weak}$, only contain an image level label per image.
An actively annotated dataset, $D_{active}$, is~newly constructed after a training session and will be concatenated to $D_{strong}$ in the next training session.
Individual data streams are maintained during training for the strongly annotated dataset and the weakly labeled dataset.
Dataflow in the network is shown in Figure~\ref{fig:dataflow}.  %MDPI: please check if it was invalid citation.
The loss for $D_{strong}$ is calculated in the same manner, as it is proposed in~\cite{i25},
where loss for the region proposal network (RPN) and the RCNN-top layer is propagated seperately.
Images in the $D_{weak}$ dataset can contribute to the classification loss in RCNN-top only when the RPN has proposed a correct region.
The loss for $D_{weak}$ will have less influence until this condition is believed to be satisfied.
After the first training session is finished, $D_{active}$ dataset is crafted from $D_{weak}$ by giving a prediction 
that is likely to contain a mass a single ground truth annotation.
Images in $D_{active}$ will be concatenated to $D_{strong}$, reducing the sparsity issue that the task originally conveyed.
The experiments show that using $D_{weak}$ images in a conservative manner helps the classifier to be detect more lesions. Training with $D_{active}$ shows an additional increase in the overall performance.
We believe that the proposed method can be adopted to general cases where strong annotations are insufficient to train the model classifier and weak labels are more~available.

\section{Materials and~Methods}
\unskip
\subsection{Datasets}

The proposed data are evaluated on the Seoul National University Bundang Hospital Breast Ultrasound (SNUBH BUS) dataset for BUS images and further tested on the Stanford Dog dataset for general images.
While the SNUBH BUS dataset has both $D_{strong}$ and $D_{weak}$ images, the~Stanford Dog dataset only contains $D_{strong}$ images.
Thus, the Stanford Dog data are manually divided into $D_{strong}$ and $D_{weak}$, where only image labels are used in images selected as $D_{weak}$.

The SNUBH dataset collected from the Seoul National University Bundang Hospital is obtained from different ultrasound systems described in~\cite{i24}, 
including Philips (ATL HDI 5000, iU22), SuperSonic Imagine (Aixplorer), and~Samsung Medison (RS80A).
The dataset contains a total of 5624 images from 2578 patients. The~$D_{strong}$ subset is comprised of 1200 images, 600 of which are benign and the other 600 of which are malignant.
We use 400 images from each class as a training set, and~200 as the test set.
$D_{weak}$ subset is comprised of 4224 images, 3291 of which are benign and the remaining 933 malignant. 
All of the image labels are proven with biopsy results, 
also meaning that the data are the cases where biopsy was needed to diagnose the patient, making classification with BUS images an even more difficult~task.

The Stanford Dog dataset is a collection of color images of 120 breeds of dogs with a total of 20,580~images,
all including class labels and bounding box coordinates.
In order to mimic the situation in BUS images, we select two similar looking middle size breeds to classify, 
the Bloodhound and the English foxhound and then converted them to grayscale images.
The number of images in each class is 187 and 157, respectively.
Each dataset is subdivided into 20 $D_{strong}$ training set, 60 test set, and~the remaining 107 and 77 images 
from Blackhound and English Foxhound, respectively, to~$D_{weak}$ dataset.
This setting enforces a situation where there are limited amount of strong annotations.
The Stanford Dog dataset is tested to demonstrate the validness of the presented method on general images.
Only a limited amount of strongly annotated images are available for training and 
the task is not straight forward, since the images are grayscale images, having room for improvement.
The dataset is available online (\url{http://vision.stanford.edu/aditya86/ImageNetDogs/}).
A summary of the number of images for both datasets is provided in Table~\ref{table: datasets}.

\begin{table}[H]
    \caption{Cardinality of SNUBH and Stanford Dog~Dataset.}
    \centering
    \begin{tabular}{p{0.1\columnwidth}p{0.3\columnwidth}p{0.1\columnwidth}p{0.1\columnwidth}p{0.1\columnwidth}p{0.1\columnwidth}p{0.1\columnwidth}p{0.1\columnwidth}}
        \toprule
        \multicolumn{2}{c}{\textbf{Dataset}} & \multicolumn{3}{c}{\textbf{SNUBH}} & \multicolumn{3}{c}{\textbf{Stanford Dog}} \\ \midrule
        \multicolumn{1}{c}{Role} & \multicolumn{1}{c}{Supervision} & \multicolumn{1}{c}{Mal.} & \multicolumn{1}{c}{Ben.} & \multicolumn{1}{c}{Total} & \multicolumn{1}{c}{Blk.} & \multicolumn{1}{c}{Eng.} & \multicolumn{1}{c}{Total} \\ \midrule
        \multicolumn{1}{c}{\multirow{2}{*}{Train}} & \multicolumn{1}{c}{Strong} & \multicolumn{1}{c}{400} & \multicolumn{1}{c}{400} & \multicolumn{1}{c}{800} & \multicolumn{1}{c}{20} & \multicolumn{1}{c}{20} & \multicolumn{1}{c}{40} \\ \cmidrule{2-8} 
        \multicolumn{1}{c}{} & \multicolumn{1}{c}{Weak} & \multicolumn{1}{c}{933} & \multicolumn{1}{c}{3291} & \multicolumn{1}{c}{4224} & \multicolumn{1}{c}{107} & \multicolumn{1}{c}{77} & \multicolumn{1}{c}{184} \\ \midrule
        \multicolumn{1}{c}{Test} & \multicolumn{1}{c}{Strong} & \multicolumn{1}{c}{200} & \multicolumn{1}{c}{200} & \multicolumn{1}{c}{400} & \multicolumn{1}{c}{60} & \multicolumn{1}{c}{60} & \multicolumn{1}{c}{120} \\ \midrule
        \multicolumn{2}{c}{Total} & \multicolumn{3}{c}{5424} & \multicolumn{3}{c}{354} \\ \bottomrule\end{tabular}

\begin{tabular}{@{}c@{}} 
\multicolumn{1}{c}{\footnotesize  Mal., Ben., Blk., Eng., denote malignant, benign, Blackhound, English  Foxhound respectively.}
\end{tabular}

    \label{table: datasets}
\end{table}
\unskip

\subsection{Training Procedure Using D\textsubscript{strong} Subset}

\begin{figure}[H]
    \centering
    \includegraphics[width=\textwidth]{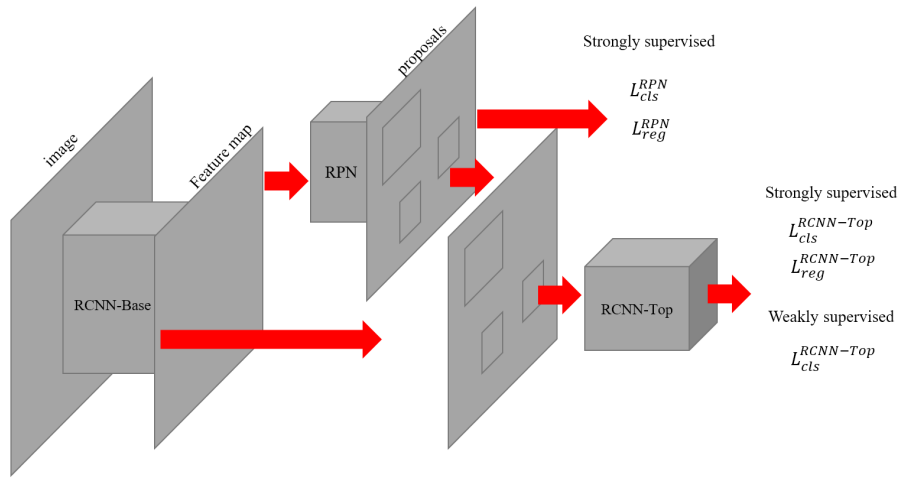}
    \caption{Illustration of the dataflow in the presented model. All of the four losses are used when training with 
    $D_{strong}$ images, while only the classification loss in RCNN-top is calculated for $D_{weak}$ images. Refer to~\cite{i25} and \ref{subsection:weak}
    for detailed methods for choosing target RoIs and calculating losses.}
    \label{fig:dataflow}
\end{figure}

The Faster-RCNN model is used for object detection tasks, which is detecting lesions in BUS images.
Faster-RCNN is a two stage object detector, where a RPN is trained to specifically perform region proposals on feature maps.
Region of interest (RoI) obtained from the RPN is then fed to the RCNN-top layer for classification and additional bounding box regression.
Bounding box information is only given by images in $D_{strong}$ subset.
This information is used for bounding box regression in both the RPN and RCNN-top, and~for foreground background classification in the RPN.
The overall dataflow is shown in Figure~\ref{fig:dataflow}.
The loss is comprised of four terms, $L_{reg}^{RPN}$, $L_{cls}^{RPN}$, $L_{reg}^{RCNN-top}$, $L_{cls}^{RCNN-top}$.
$L_{reg}^{RPN}$ and $L_{reg}^{RCNN-top}$, which are regression losses for the RPN and the RCNN-top, respectively, are obtained by calculating the smooth l1 loss between the ground truth box and the predicted box coordinates.
$L_{cls}^{RPN}$ and $L_{cls}^{RCNN-top}$, which are classification losses for the RPN and the RCNN-top, respectively, are obtained by calculating the crossentropy loss between the ground truth label and the predicted label.
Corresponding ground truth label and coordinates are assigned when the intersection over union (IoU) between the boxes are over 0.5.
Details of calculating the four terms remain same as the method that was proposed in~\cite{i25}.

\unskip

\subsection{Training Procedure Using D\textsubscript{weak} Subset}
\label{subsection:weak}

Without bounding box annotations, bounding box regression or foreground background classification can be performed.
Thus, images in the $D_{weak}$ dataset can only aid the classification procedure in the RCNN-top section.
We must have a strategy for giving labels to RoIs proposed by the RPN in order to use $D_{weak}$ images.
Although there is no complete way for figuring out the labels of each RoIs, it is known that given an image label,
there is at least one mass that should be labeled as the image label. 
We are able to infer the most probable RoI that should be labeled by rewriting the model with random variables.
Let $X_{roi}$ be indicator random variables that map all RoIs to their ground truth (background, benign, malignant) and $G$ be the set of all RoIs in an image.
Set G is obtained as an output of the RPN. RoIs in G are considered to contain distinct objects after the non-maximum suppression (NMS) post-processing. 
NMS eliminates RoIs that overlap with an IoU over 0.5. The~RoI with the higher foreground score is kept among the two RoIs.
The relationship of the values is defined as $malignant>benign>background$. 
\[ Y = \underset{i \in G}{\max}( X_{roi_{1}}, X_{roi_{2}}, X_{roi_{3}},...) \]

Thus, $Y$ represents the label of an image, since a single malignant lesion would make an image label malignant,
and a single benign lesion would make the image label benign if there are no other malignant lesions.
Subsequently, the most probable mass to be labeled given the image label can be written, as~follows,
\[ \underset{i \in G}{\argmax} \ p(X_{roi_{i}}=labelY=label) \equiv \underset{i \in G}{\argmax} \ p(X_{roi_{i}}=label). \]

Because $Y$ is a max of all RoI labels, conditioning the probability with $Y=label$ gives no information 
if the probability in question is that of $X$ having the same label.
Thus, it is optimal to choose the RoI with the highest probability of containing the labeled object.
Let $\hat{X}$ denote the mapping between the proposed RoI and the predicted label by the RCNN-top layer.
Since $\hat{X}$ is trained directly by the cross entropy loss with $X$ when using the $D_{strong}$ dataset, 
$\hat{X}$ can be used as an alternative of $X$ if suitably trained.
Therefore, we label the RoI with the highest image label score after running through RCNN-top, to~be the train target in the RCNN-top section 
and then calculate the loss for a single $D_{weak}$ image, as follows,
\[ L_{cls}^{RCNN-top} = crossentropy( \underset{i \in G}{\max} \ p(\hat{X_{i}}),p(label)). \]

However, $\hat{X}$ would not be able to replace $X$ in the early stages of training. 
Hence, we introduce a controlled weight for $L_{cls}^{RCNN-top}$, so-called $\alpha$.
We increase $\alpha$ from a 0.01 as the training progresses and the manner of this increase can vary. 
The weight $\alpha$ for $L_{cls}^{RCNN-top}$ was selected among the following~candidates:
\[ \alpha = 1 \]
\[ \alpha = 1 - 0.99 \: (0.9^{step / 2000}) \]
\[ \alpha = 0.01 + 0.99 \: (step/totalsteps) \]
\begin{equation} \alpha = 0.01 + 0.99 \: (step/totalsteps)^5 \label{exp5} \end{equation}
\begin{equation} \alpha = 0.01 + 0.99 \: (step/totalsteps)^{16}\label{exp16} \end{equation}
\begin{equation} \alpha = 0.01 + 0.99 \: (step/totalsteps)^{32} \label{exp32} .\end{equation}

\begin{figure}[H]
    \centering
    \includegraphics[width=0.85\textwidth]{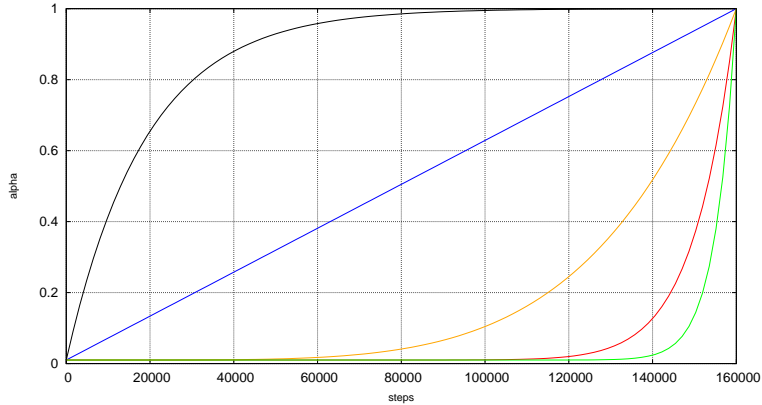}
    \caption{Illustration for the tendancy of alpha when the step size is 160000.
    The black plot shows a log-like increase, namely inverse exponential, in~$\alpha$, which converges to 1 quickly.
    The blue plot is a linear increase of $\alpha$.
    \eqref{exp5}, \eqref{exp16}, \eqref{exp32} are the conservative increase of $\alpha$ during the training phase, namely polynomial increase, which relates to the orange plot, red plot, and green plot, respectively.\label{fig:wsgraph}}
\end{figure}

Changes of $\alpha$ following the training steps are visualized in Figure~\ref{fig:wsgraph}.
The usage of $L_{cls}^{RCNN-top}$ is considered to be more conservative as the equation number increases.
The $D_{weak}$ and $D_{strong}$ dataset is concurrently used and the calculated loss from each image is summed, as follows,

\[L_{final} = L_{strong} + \alpha \: L_{weak},\]
\[L_{strong} = L_{cls}^{RPN} + L_{reg}^{RPN} + L_{cls}^{RCNN-top} + L_{reg}^{RCNN-top},\]
\[L_{weak} = L_{cls}^{RCNN-top}.\]

\unskip

\subsection{D\textsubscript{active} Construction with D\textsubscript{weak} Test~Results}
    
\begin{figure}[H]
    \centering
    \includegraphics[width=65mm]{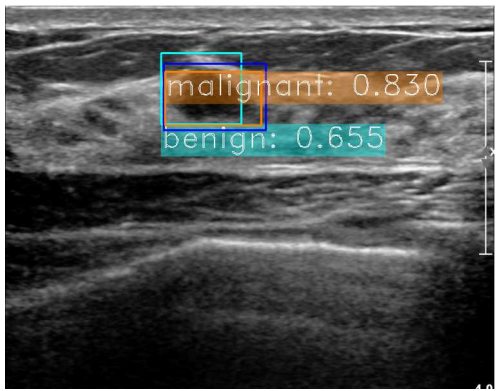}
    \caption{Example of a double prediction case in Breast ultrasound (BUS) images. The~bounding box in blue represents the ground truth for a benign mass.
    Predicted boxes are colored in orange and cyan for malignant and benign predictions, respectively.\label{fig:fig2}}
\end{figure}

$D_{active}$ is a dataset that we create with the $D_{weak}$ dataset by adding annotations that are generated from the initial model after a training is finished.
$D_{active}$ dataset can aid the $D_{strong}$ dataset, since images in $D_{strong}$ are assumed to be insufficient in this problem setting.
Predicted bounding boxes and predictions are not reliable in itself, 
which requires the cautious selection of images to include.
Verifying whether a predicted bounding box contains an object or not is the main issue.
The double prediction problem can be a benefit for solving this problem.
Double prediction is the case when two different predictions are made for a single object, as seen in Figure~\ref{fig:fig2}.
Objects in double predicted boxes are more likely to contain an object than other predicted boxes, 
since it was predicted to contain a lesion twice.
We can generate a strong annotation by selecting the correct labeled box of the two predicted boxes.
The image level label is used to pick the correct bounding box among the two uncertain predictions.

All of the images in $D_{weak}$ are tested through the trained model, generating multiple bounding boxes with labels for each image.
We iterate through the boxes in an image to check whether there is a double prediction based on the PASCAL VOC criteria, which defines boxes to be overlapping when their IoU is higher than 0.5.
If multiple double prediction pairs exist for an image, we choose the pair with the higher IoU.
Once a pair is selected for an image, we annotate the image with the bounding box that hold the original image label.
Newly annotated images will contain a bias towards benign, since $D_{weak}$ is biased.
Thus, we only choose malignant images to add to the $D_{active}$ dataset to compensate this bias,
and also due to the medical imaging setting where a failure to detect a malignant lesion is critical.
The newly generated $D_{active}$ dataset is used in the same manner as the $D_{strong}$ dataset, since they can now produce the same type of losses.

\subsection{Faster-RCNN Hyperparameters and Model~Details}
\label{subsection:details}

We use the PASCAL VOC pre-trained VGG-16~\cite{m1} as the backbone for generating feature maps, 
only fine-tuning the final layers higher than conv3\_1, which is the method used by the original Faster-RCNN~\cite{i25}.
The RPN’s regression and classification network was modified to use 3$\times$3 convolution instead of 1$\times$1 for better detection of objects.
We reduced the size of the fully connected layer in the RCNN-top to 2048 to prevent overfitting.
The $D_{strong}$ dataset was augmented by horizontal flipping, which increases the number of images, 
and, by random brightness, contrast adjustments given to images, which preserves the number of images.
Steps are used to check the training progress, since epochs cannot be calculated when using two datasets with different sizes.
One step corresponds to using a single batch from each datasets.
Th Adam optimizer was used for optimization, with~a configuration of batch size 1 for each dataset.
Negative sampling for background RoI was performed when training $D_{weak}$ images, since the choosing a RoI with the image label for $D_{weak}$ images makes the distribution of RoIs unbalanced. 
The least scoring box was labeled as background for the RCNN-top to calculate.
Class weights were also given for $D_{weak}$ losses, since the dataset has a bias towards benign.
All of the details and code of the model will be available online (\url{https://github.com/YeolJ00/faster-rcnn-pytorch}) for research purposes.

\section{Results}\unskip
\subsection{Evaluation~Specifications}

In this study, a~model generates multiple bounding boxes for an image. 
Each detection is considered to be a true positive (TP) if the classified label of the detection matches the target GT class, and the IoU between the predicted bounding box and the target GT is higher than 0.5. Otherwise, \mbox{it is regarded} as a false positive (FP).
We evaluate the performance of the model with the test images in SNUBH and Stanford Dog dataset through some measures 
such as correct localization (CorLoc), and~fraction of lesion detected.

CorLoc is defined as the ratio of correctly classified and localized images.
A correctly classified image is an image that contains a TP detection in its predicted boxes.
Although mean average precision is widely used for general deep learning models, 
CorLoc is more applicable the BUS case, since detecting a positive mass is critical in medical imaging.
Additionally, only a single mass in an image is labeled as GT, while there could be other possible unlabeled masses,
thus FP detections might actually contain masses.
The fraction of lesion detected is the measure for localization performance, 
which is obtained by the ratio of images that have a bounding box that overlaps with its GT box.

\subsection{Experiments for Controlling the Effect of Weakly Annotated Images in SNUBH Dataset}

Table~\ref{table:alphas} presents the quantitative result of the experiments.
The experiments are conducted on a total of 160,000 training steps, 
and all of the hyperparameters except $\alpha$ are equally applied.
\mbox{It is found} that a model does not perform well when $\alpha$ is a constant value or 
increased with an inverse exponential function.
We believe that the value was too high in the early stages of training.
$L_{weak}$ was not penalized enough before RPN was trained enough to provide valid RoI proposals, 
which gives an incorrect loss for the classifier.
Based on this idea, we compared more conservative functions for increasing $\alpha$.
We can see that all of the subsequent methods demonstrate an improvement both in CorLoc and~the fraction of lesion detected.
The fraction of lesion detected is the fraction of ground truth lesions that were given a bounding box.
Performance tends to increase as $\alpha$ is maintained low during most of the training phase, 
and the model exhibited the best result when $\alpha$ followed \eqref{exp16}.
24\% point CorLoc increase and a 20\% point fraction of lesion detected increase was shown as compared to the model without controlled weight.
A slight loss of performance was shown when $\alpha$ follows \eqref{exp32}.
We believe this is due to a drastically increasing $\alpha$ for the case when the total step is 160,000, 
making the loss increase faster than the optimization step.
Additionally, weakly annotated data was fully used only for a small number of steps in \eqref{exp32}.
\begin{table}[H]
    \caption{Results showing variants of controled weight $\alpha$ with the SNUBH BUS~dataset.
}
    \centering
    \begin{tabular}{ccc}
    \toprule
    \multicolumn{1}{c}{$\boldsymbol{\alpha}$ \textbf{Control Schedule}} & \multicolumn{1}{c}{\textbf{CorLoc [\%]}} & \multicolumn{1}{c}{\begin{tabular}[c]{@{}c@{}}\textbf{Fraction of}\\ \textbf{Lesion Detected [\%]}\end{tabular}} \\ \midrule
    \multicolumn{1}{c}{constant} & \multicolumn{1}{c}{41.75} & \multicolumn{1}{c}{56.00} \\ \midrule
    \multicolumn{1}{c}{inverse exponential} & \multicolumn{1}{c}{49.75} & \multicolumn{1}{c}{69.25} \\ \midrule
    \multicolumn{1}{c}{linear} & \multicolumn{1}{c}{60.25} & \multicolumn{1}{c}{70.50} \\ \midrule
    \multicolumn{1}{c}{polynomial \eqref{exp5}} & \multicolumn{1}{c}{58.75} & \multicolumn{1}{c}{67.00} \\ \midrule
    \multicolumn{1}{c}{\begin{tabular}[c]{@{}c@{}}proposed: \\ polynomial \eqref{exp16}\end{tabular}} & \multicolumn{1}{c}{\textbf{65.75}} & \multicolumn{1}{c}{\textbf{76.00}} \\ \midrule
    \multicolumn{1}{c}{polynomial \eqref{exp32}} & \multicolumn{1}{c}{63.00} & \multicolumn{1}{c}{74.50} \\ \bottomrule
   \end{tabular}

\begin{tabular}{@{}c@{}} 
\multicolumn{1}{p{\textwidth -.88in}}{\footnotesize Correct localization (CorLoc)} and fraction on lesion detected according to the manner of how $\alpha$ is increase. CorLoc measures both classification and localization performance while fraction of lesion detected only measures the localization performance. Detailed equations are presented in Section~\ref{subsection:weak}.  %MDPI: please add the explaination for the bold.

\end{tabular}

    \label{table:alphas}
\end{table}
\unskip

\begin{figure}[H]
    \centering
    \includegraphics[width=\textwidth]{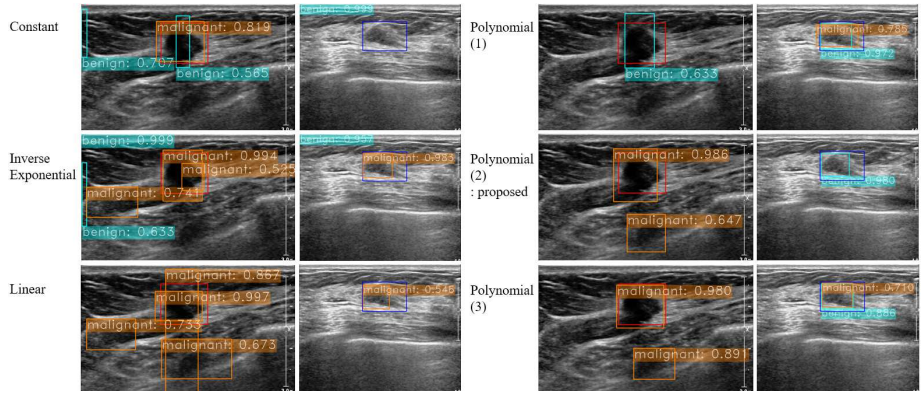}
    \caption{Qualitative results for controlling $\alpha$. Bounding boxes colored red/blue are ground truth boxes for malignant/benign masses.
    Bounding boxes colored orange/cyan are predictions for malignant/benign masses. Two cases are presented for each method.\label{fig:alphas}}
\end{figure}

Qualitative results for controlling $\alpha$ are shown in Figure~\ref{fig:alphas}.
The proposed schedule for $\alpha$ shows both solid localization of objects and classification of bounding box proposals.
Figure~\ref{fig:alphas} also shows a false positive detection for the proposed method, 
yet the false positive detection has a relatively low score of being malignant when compared to the method following \eqref{exp32}.
\subsection{Experiments for Active Learning on SNUBH~Dataset}

Quantitative results for active learning experiment is shown in Table~\ref{table:active}.
$D_{active}$ constructed from the model trained with the proposed $\alpha$ weight \eqref{exp16} consists of 238 malignant images.
Active learning aims to extend the $D_{strong}$ dataset, which is the primary dataset that trains the model.
Performing active learning gives a 2.75\% increase in CorLoc measure and a 3.75\% increase in the fraction of lesion detected measure.
Both classification and localization performance has increased.
\begin{table}[H]
    \caption{Results showing the effect of active learning in SNUBH~dataset.}
    \centering
    \begin{tabular}{ccc}
    \toprule
    \multicolumn{1}{c}{\begin{tabular}[c]{@{}c@{}}\textbf{Active Learing}\end{tabular}} & \multicolumn{1}{c}{\textbf{CorLoc [\%]}} & \multicolumn{1}{c}{\begin{tabular}[c]{@{}c@{}}\textbf{Fraction of}\\ \textbf{Lesion Detected [\%]}\end{tabular}} \\ \midrule
    \multicolumn{1}{c}{before} & \multicolumn{1}{c}{65.75} & \multicolumn{1}{c}{76.00} \\ 
    \multicolumn{1}{c}{after} & \multicolumn{1}{c}{68.50} & \multicolumn{1}{c}{79.75} \\ \bottomrule
 
    \end{tabular}

\begin{tabular}{@{}c@{}} 
\multicolumn{1}{c}{\footnotesize CorLoc and Fraction of lesion detected before and after active learning  is presented.}
\end{tabular}

    \label{table:active}
\end{table}

\begin{figure}[H]
    \centering
    \includegraphics[width=\textwidth]{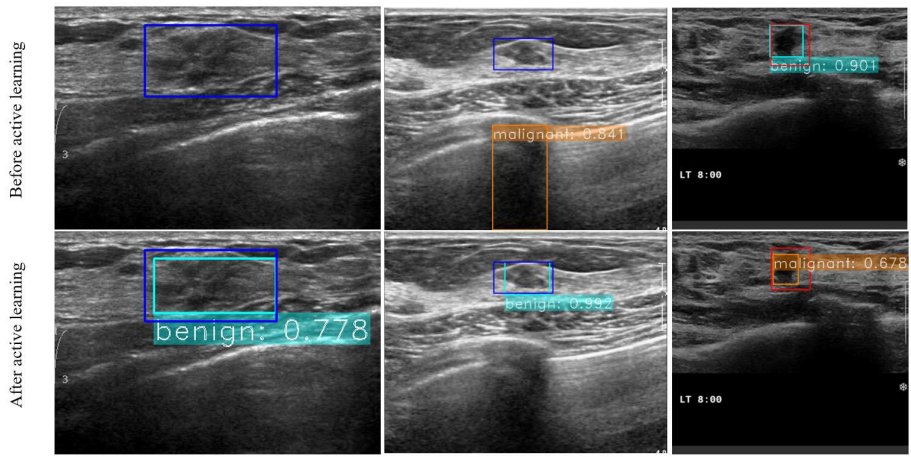}
    \caption{Qualitative results for controlling $\alpha$. Bounding boxes colored red/blue are ground truth boxes for malignant/benign masses.
    Bounding boxes colored orange/cyan are predictions for malignant/benign masses. Boxes on the left are the results before active learning,
    and the right side shows the same predictions made for the images after active learning.\label{fig:active}}
\end{figure}

Figure~\ref{fig:active} presents the qualitative results.
Some masses that were difficult to detect or classify were given the correct predictions after
training with $D_{active}$.
Both localization and classification performance are enhanced.

\unskip

\subsection{Experiments on Comparable Object~Detectors}

The proposed model was compared with other object detectors in~\cite{i24} and~\cite{i25}.
A vanilla Faster-RCNN model was trained with $D_{strong}$ images while using the specifications introduced in~\cite{i24}.
The~Faster-RCNN based model in~\cite{i25} is a model that uses weakly annotated images jointly with strong, bounding box annotations.
Thus, we were able to reconstruct the model to train with the SNUBH dataset.
Implementations of the models are provided online (\url{https://github.com/YeolJ00/faster-rcnn-pytorch}).
Table~\ref{table:general} shows the results.

\begin{table}[H]
    \caption{Results for various object~detectors.}
    \centering
    \begin{tabular}{ccc}
    \toprule
    \multicolumn{1}{c}{\textbf{Detectors}} & \multicolumn{1}{c}{\textbf{CorLoc [\%]}} & \multicolumn{1}{c}{\begin{tabular}[c]{@{}c@{}}\textbf{Fraction of}\\ \textbf{Objects Detected [\%]}\end{tabular}} \\ \midrule
    \multicolumn{1}{c}{Vanilla Faster-RCNN~\cite{i24}} & \multicolumn{1}{c}{42.50} & \multicolumn{1}{c}{57.50} \\ \midrule
    \multicolumn{1}{c}{\begin{tabular}[c]{@{}c@{}}Weakly supervised\\ Faster-RCNN~\cite{i25}\end{tabular}} & \multicolumn{1}{c}{33.75} & \multicolumn{1}{c}{59.00} \\ \midrule
    \multicolumn{1}{c}{proposed} & \multicolumn{1}{c}{68.50} & \multicolumn{1}{c}{79.75} \\ \bottomrule
  
    \end{tabular}

\begin{tabular}{ccc}
\multicolumn{1}{c}{\footnotesize CorLoc, Fraction of objects detected is shown for different object detectors.}
\end{tabular}

    \label{table:general}
\end{table}
\unskip

\subsection{Experiments on Stanford Dog~dataset}

Experiments for controlled weight and active learning was performed with the Stanford Dog~dataset.

\begin{figure}[H]
    \centering
    \includegraphics[width=\textwidth]{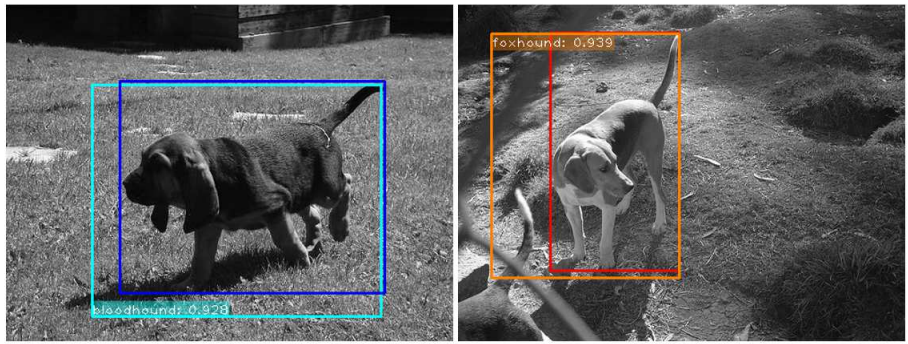}
    \caption{Prediction results from the Stanford Dog images. Image on the right and left are predictions for a Blackhound image and a English Foxhound image, respectively.\label{fig:dogs}}
\end{figure}

The results for controlling $\alpha$ and active learning are summarized in Tables~\ref{table:dogalphas} and \ref{table:dogactive} respectively.
Little increase in CorLoc was shown for the proposed $\alpha$ control method.
We believe that the reason behind the negligible performance increase for the proposed $\alpha$ control method 
is due to the big bounding box proportion in the images.
This enables the RPN to propose correct bounding boxes at an earlier stage of the training, which means that the loss is less likely to be lead to a local minimum.
Acitve learning added 23 images to the strong annotated dataset, 10 Blackhound boxes, and 13 English Foxhound boxes.
We included images from both classes, since this is not a medical imaging task where a detecting a certain class is preferred.
Performing active learning on the trained model shows a slight decrease in CorLoc measures, 
which is a measure that ignores FP predcitions.
However, the widely used measure of performance for object detection tasks is mAP, which increased by 17.46\% point after active learning.
The increase in strong annotations has reduced false positive predictions, significantly increasing the precision of the model.
Model performance does not vary much due to the generally high performance.
The prediction result samples can be viewed in Figure~\ref{fig:dogs}.

\begin{table}[H]
    \caption{Results showing variants of controled weight $\alpha$ with the Stanford Dog~dataset.}
    \centering
    \begin{tabular}{ccc}
    \toprule
    \multicolumn{1}{c}{$\boldsymbol{\alpha}$ \textbf{Increase Method}} & \multicolumn{1}{c}{\textbf{CorLoc [\%]}} & \multicolumn{1}{c}{\begin{tabular}[c]{@{}c@{}}\textbf{Fraction of}\\ \textbf{Objects Detected [\%]}\end{tabular}} \\ \midrule
    \multicolumn{1}{c}{constant } & \multicolumn{1}{c}{83.33} & \multicolumn{1}{c}{86.67} \\ \midrule
    \multicolumn{1}{c}{inverse exponential} & \multicolumn{1}{c}{85.83} & \multicolumn{1}{c}{87.50} \\ \midrule
    \multicolumn{1}{c}{linear} & \multicolumn{1}{c}{83.33} & \multicolumn{1}{c}{87.50} \\ \midrule
    \multicolumn{1}{c}{polynomial \eqref{exp5}} & \multicolumn{1}{c}{79.17} & \multicolumn{1}{c}{84.17} \\ \midrule
    \multicolumn{1}{c}{\begin{tabular}[c]{@{}c@{}}proposed: \\ polynomial \eqref{exp16}\end{tabular}} & \multicolumn{1}{c}{\textbf{87.50}} & \multicolumn{1}{c}{\textbf{89.17}} \\ \midrule
    \multicolumn{1}{c}{polynomial \eqref{exp32}} & \multicolumn{1}{c}{81.67} & \multicolumn{1}{c}{87.50} \\ \bottomrule
   
    \end{tabular}

\begin{tabular}{@{}c@{}} 
\multicolumn{1}{p{\textwidth -.88in}}{\footnotesize CorLoc and fraction on objects detected according to the manner of 
    how $\alpha$ is increase. Detailed equations remain same as the test  with SNUBH BUS dataset.}
\end{tabular}

    \label{table:dogalphas}
\end{table}
\unskip

\begin{table}[H]
    \caption{Results showing the effect of active learning in the Stanford Dog~dataset.}
    \centering
    \begin{tabular}{cccc}
    \toprule
    \multicolumn{1}{c}{\begin{tabular}[c]{@{}c@{}}\textbf{Active}\\ \textbf{Learing}\end{tabular}} & \multicolumn{1}{c}{\textbf{CorLoc [\%]}} & \multicolumn{1}{c}{\begin{tabular}[c]{@{}c@{}}\textbf{Fraction of Lesion Detected [\%]}\end{tabular}} & \multicolumn{1}{c}{\textbf{mAP [\%]}} \\ \midrule
    \multicolumn{1}{c}{before} & \multicolumn{1}{c}{87.50} & \multicolumn{1}{c}{89.17} & \multicolumn{1}{c}{36.84} \\  
    \multicolumn{1}{c}{after} & \multicolumn{1}{c}{84.17} & \multicolumn{1}{c}{87.50} & \multicolumn{1}{c}{54.30} \\ \bottomrule
    \end{tabular}
\begin{tabular}{@{}c@{}} 
\multicolumn{1}{p{\textwidth -.88in}}{\footnotesize lCorLoc, fraction of lesion detected, and~mean average precision (mAP) before and after active learning is~presented.}
\end{tabular}

    \label{table:dogactive}
\end{table}
\unskip

\unskip

\section{Conclusions and~Discussion}
We propose an applicable mechanism for utilizing weakly annotated images for object detection models in a setting 
where bounding box information is insufficient for achieving high classification performance.
The proposed method enables a successful increase of the size of strong annotations by safely assigning bounding box predictions as ground truth.
The method is applied to the primary task of detecting masses in BUS and tested on the Stanford Dog dataset to verify generality.
A comparison with different variants of the method supports the reasoning behind the manner of controlling the influence of weakly annotated images.
We notice that maintaining the loss from weakly annotated images at a low level until the RPN proposes bounding boxes containing objects 
guides the model to have a higher classification capability.
Additionally, we set specific configurations for the active learning scheme, which can be a risky work, since there is no way to confirm the correct assigning of GT bounding boxes.
The results show that it can enhance classification performance if it was an issue.

For our future work, we plan to extend the proposed method to autonomously detect whether if 
the RPN is proposing bounding boxes containing objects and control the weight, which was originally increased following a fixed schedule.
This will increase the generality of the method, since the point of RPN convergence may vary depending on the size and detection difficulty of a dataset.
We believe that the proposed method can be applied to typical cases of medical imaging tasks where strong annotations are costly and weakly labeled data are relatively easy to obtain from the diagnosis procedure.

%%%%%%%%%%%%%%%%%%%%%%%%%%%%%%%%%%%%%%%%%%
\vspace{6pt} 


\begin{thebibliography}{999}
\providecommand{\natexlab}[1]{#1}

\bibitem[Cheng \em{et~al.}(2010)Cheng, Shan, Ju, Guo, and Zhang]{i11}
Cheng, H.; Shan, J.; Ju, W.; Guo, Y.; Zhang, L.  %MDPI: ref 1 and 8 are same.
\newblock Automated breast cancer detection and classification using ultrasound
  images: A survey.
\newblock {\em Pattern Recognit.} {\bf 2010}, {\em 43},~299--317, doi:10.1016/j.patcog.2009.05.012.

\bibitem[Cheng \em{et~al.}(2006)Cheng, Shi, Min, Hu, Cai, and Du]{i12}
Cheng, H.D.; Shi, X.; Min, R.; Hu, L.; Cai, X.; Du, H.
\newblock Approaches for automated detection and classification of masses in
  mammograms.
\newblock {\em Pattern Recognit.} {\bf 2006}, {\em 39},~646--668.

\bibitem[Stavros \em{et~al.}(1995)Stavros, Thickman, Rapp, Dennis, Parker, and
  Sisney]{i13}
Stavros, A.T.; Thickman, D.; Rapp, C.L.; Dennis, M.A.; Parker, S.H.; Sisney,
  G.A.
\newblock Solid breast nodules: Use of sonography to distinguish between benign
  and malignant lesions.
\newblock {\em Radiology} {\bf 1995}, {\em 196},~123--134.

\bibitem[Drukker \em{et~al.}(2008)Drukker, Gruszauskas, Sennett, and
  Giger]{i14}
Drukker, K.; Gruszauskas, N.P.; Sennett, C.A.; Giger, M.L.
\newblock Breast US computer-aided diagnosis workstation: Performance with a
  large clinical diagnostic population.
\newblock {\em Radiology} {\bf 2008}, {\em 248},~392--397.

\bibitem[Ragesh \em{et~al.}(2011)Ragesh, Anil, and Rajesh]{i15}
Ragesh, N.; Anil, A.; Rajesh, R.
\newblock Digital image denoising in medical ultrasound images: A survey.
\newblock  In~{Proceedings of the }Icgst Aiml-11 Conference, Dubai, UAE,  12 April 2011; Volume~12, p.~14.

\bibitem[Madjar(2010)]{i16}
Madjar, H.
\newblock Role of breast ultrasound for the detection and differentiation of
  breast lesions.
\newblock {\em Breast Care} {\bf 2010}, {\em 5},~109--114.

\bibitem[Hansen \em{et~al.}(2008)Hansen, Huttebrauker, Schasse, Wilkening,
  Ermert, Hollenhorst, Heuser, and Schulte-Altedorneburg]{i17}
Hansen, C.; Huttebrauker, N.; Schasse, A.; Wilkening, W.; Ermert, H.;
  Hollenhorst, M.; Heuser, L.; Schulte-Altedorneburg, G.
\newblock Ultrasound breast imaging using full angle spatial compounding:
  In-vivo results.
\newblock  In~{Proceedings of the }2008 IEEE Ultrasonics Symposium,  Beijing, China, 2--5 November 2008; pp.~54--57.

\bibitem[Pons \em{et~al.}(2014)Pons, Mart{\'\i}, Ganau, Sent{\'\i}s, and
  Mart{\'\i}]{i22}
Pons, G.; Mart{\'\i}, R.; Ganau, S.; Sent{\'\i}s, M.; Mart{\'\i}, J.
\newblock Computerized detection of breast lesions using deformable part models
  in ultrasound images.
\newblock {\em Ultrasound Med. Biol.} {\bf 2014}, {\em
  40},~2252--2264.

\bibitem[Yap \em{et~al.}(2018)Yap, Goyal, Osman, Mart{\'\i}, Denton, Juette,
  and Zwiggelaar]{i23}
Yap, M.H.; Goyal, M.; Osman, F.M.; Mart{\'\i}, R.; Denton, E.; Juette, A.;
  Zwiggelaar, R.
\newblock Breast ultrasound lesions recognition: End-to-end deep learning
  approaches.
\newblock {\em J. Med. Imaging} {\bf 2018}, {\em 6},~011007.

\bibitem[Shin \em{et~al.}(2018)Shin, Lee, Yun, Kim, and Lee]{i24}
Shin, S.Y.; Lee, S.; Yun, I.D.; Kim, S.M.; Lee, K.M.
\newblock Joint weakly and semi-supervised deep learning for localization and
  classification of masses in breast ultrasound images.
\newblock {\em IEEE Trans. Med. Imaging} {\bf 2018}, {\em
  38},~762--774.

\bibitem[Ren \em{et~al.}(2015)Ren, He, Girshick, and Sun]{i25}
Ren, S.; He, K.; Girshick, R.; Sun, J.
\newblock Faster r-cnn: Towards real-time object detection with region proposal
  networks.
\newblock In \emph{Advances in Neural Information Processing Systems};  2015; Montreal, Canada %Please add the location of the publisher.
 pp.
  91--99.

\bibitem[Wang \em{et~al.}(2015)Wang, Liu, Li, and Lu]{i31}
Wang, Y.; Liu, J.; Li, Y.; Lu, H.
\newblock Semi-and weakly-supervised semantic segmentation with deep
  convolutional neural networks.
\newblock   In Proceedings of the 23rd ACM international conference on Multimedia,
  Brisbane, Australia, 13 October 2015; pp. 1223--1226.

\bibitem[Zhang \em{et~al.}(2018)Zhang, Gopalakrishnan, Lu, Summers, Moss, and
  Yao]{i34}
Zhang, L.; Gopalakrishnan, V.; Lu, L.; Summers, R.M.; Moss, J.; Yao, J.
\newblock Self-learning to detect and segment cysts in lung CT images without
  manual annotation.
\newblock  In~{Proceedings of the }2018 IEEE 15th International Symposium on Biomedical Imaging (ISBI
  2018),   Washington, DC, USA,  4--7 April 2018; pp. 1100--1103.

\bibitem[Simonyan and Zisserman(2014)]{m1}
Simonyan, K.; Zisserman, A.
\newblock Very deep convolutional networks for large-scale image recognition.
\newblock {\em arXiv} {\bf 2014},   arXiv:1409.1556.
\end{thebibliography}
\end{document}